\documentclass[10pt,twocolumn,letterpaper]{article}

\usepackage{iccv}
\usepackage{times}
\usepackage{epsfig}
\usepackage{graphicx}
\usepackage{amsmath}
\usepackage{amssymb}
\usepackage{booktabs}
\usepackage{multirow}
\usepackage{subcaption}
\usepackage{comment}
\usepackage{makecell}
\usepackage[accsupp]{axessibility}
\usepackage[table]{xcolor}

\usepackage{microtype}

\usepackage[pagebackref=true,breaklinks=true,letterpaper=true,colorlinks,bookmarks=false]{hyperref}

\usepackage[capitalize]{cleveref}
\crefname{section}{Sec.}{Secs.}
\Crefname{section}{Section}{Sections}
\Crefname{table}{Table}{Tables}
\crefname{table}{Tab.}{Tabs.}

\iccvfinalcopy 

\ificcvfinal\fi

\makeatletter
\def\thanks#1{\protected@xdef\@thanks{\@thanks
        \protect\footnotetext{#1}}}
\makeatother

\begin{document}

\title{Towards Real-World Burst Image Super-Resolution: Benchmark and Method}

\author{
Pengxu Wei$^1$ \quad Yujing Sun$^2$ \quad Xingbei Guo$^1$ \quad Chang Liu$^3$ \quad Jie Chen$^2$ \\ Xiangyang Ji$^3$ \quad Liang Lin$^1$\thanks{\emph{*Corresponding author: Liang Lin, Jie Chen}} \and
$^{1}$ Sun Yat-sen University \\
$^{2}$ School of Electronic and Computer Engineering, Peking University, Shenzhen, China \\
$^{3}$ Tsinghua University \\
\small{\tt{\{weipx3, liguanbin\}@mail.sysu.edu.cn, yujingsun1999@gmail.com, guoxb7@mail2.sysu.edu.cn}}\\
\small{\tt{\{liuchang2022,xyji\}@tsinghua.edu.cn, chenj@pcl.ac.cn, linliang@ieee.org}}
}

\maketitle

\begin{abstract}
Despite substantial advances, single-image super-resolution (SISR) is always in a dilemma to reconstruct high-quality images with limited information from one input image, especially in realistic scenarios. In this paper, we establish a large-scale real-world burst super-resolution dataset, i.e., RealBSR, to explore the faithful reconstruction of image details from multiple frames. Furthermore, we introduce a Federated Burst Affinity network (FBAnet) to investigate non-trivial pixel-wise displacements among images under real-world image degradation. Specifically, rather than using pixel-wise alignment, our FBAnet employs a simple homography alignment from a structural geometry aspect and a Federated Affinity Fusion (FAF) strategy to aggregate the complementary information among frames. 
Those fused informative representations are fed to a Transformer-based module of burst representation decoding. Besides, we have conducted extensive experiments on two versions of our datasets, i.e., RealBSR-RAW and RealBSR-RGB. Experimental results demonstrate that our FBAnet outperforms existing state-of-the-art burst SR methods and also achieves visually-pleasant SR image predictions with model details. Our dataset, codes, and models are publicly available at \href{https://github.com/yjsunnn/FBANet}{https://github.com/yjsunnn/FBANet}.  
\end{abstract}

\section{Introduction}\label{sec:intro}


As a fundamental research topic, Super-Resolution (SR) attracts long-standing substantial interest, which targets high-resolution (HR) image reconstruction from a single or a sequence of low-resolution (LR) observations. 
In recent years, we have witnessed the prosperity of Single Image Super-Resolution (SISR), \eg, SRCNN~\cite{SRCNN}, EDSR~\cite{EDSR}, SRGAN~\cite{SRGAN}, RDN~\cite{RDN} and ESRGAN~\cite{ESRGAN}. 
Nevertheless, SISR intrinsically suffers from a limited capacity of restoring details from only one LR image, typically yielding over-smooth LR predictions, especially for large-scale factors. 
With real detailed sub-pixel displacement information, Multi-Frame Super-Resolution (MFSR)~\cite{Handheld, DBSR, MFIR, EBSR, BSRT} provides a promising potential to reconstruct the high-quality image from multiple LR counterparts, which is valuable for many sensitive realistic applications, \eg, medical imaging, and remote satellite sensing.

\begin{figure}[t]
     \centering
     \includegraphics[width=0.48\textwidth]{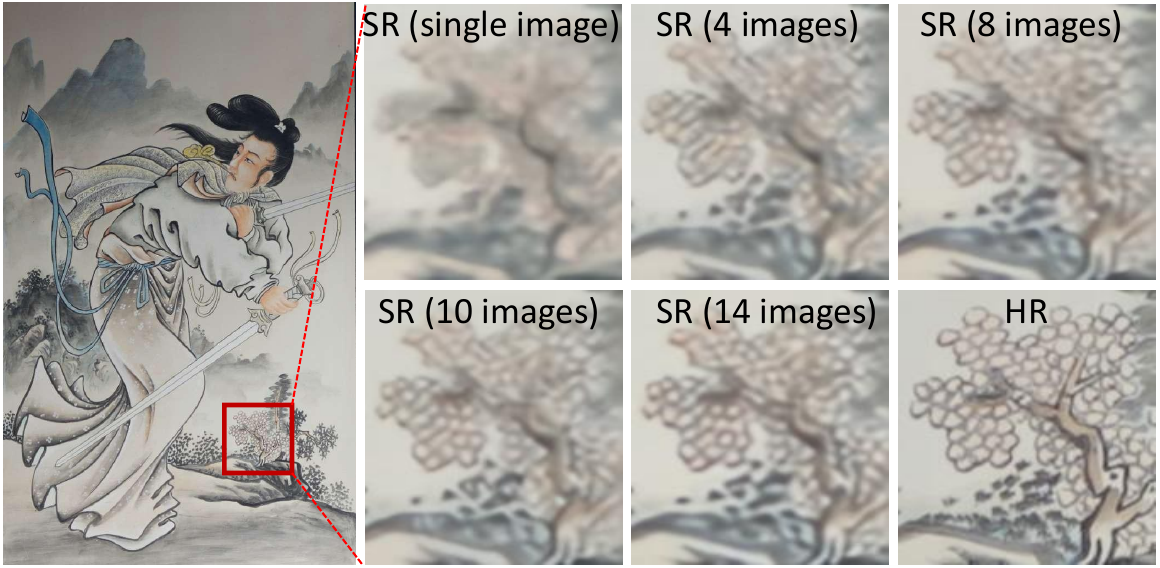}
     \vspace{-18pt}
    \caption{SR predictions with different numbers of burst image inputs in our RealBSR dataset, where more burst inputs facilitate more accurate reconstruction of image details.}
    \vspace{-8pt}
     \label{fig:multiframe_sr}
\end{figure}

After the pioneering work~\cite{tsai1984multiframe} of Tsai and Huang in 1984, the research on MFSR has not achieved as tremendous progress as SISR. 
%
Typically, they are overwhelmed by two challenges: 
1) the difficulty of fusing multiple LR inputs, which especially is aggravated for real-world data;  
2) the limitation of artificially-synthesized data, accounting for a poor generalization for real-world scenarios; 
To address those challenges, a recent work~\cite{DBSR} has made seminal contributions to the first real-world burst SR dataset benchmark, BurstSR, and a novel architecture, DBSR. 
\textls[-2]{Subsequently, MFIR proposes a deep reparametrization to reformulate the classical MAP objective in a deep feature space~\cite{MFIR}. BIPNet~\cite{BIPNet} introduces a set of pseudo-burst features for information exchange among multiple burst frames. BSRT~\cite{BSRT} employs a pyramid flow-guided deformable convolution network in a Transformer architecture.} 


Despite great progress achieved, two aspects still need to be revisited.  
\textbf{\emph{1) Method:}} 
Align-fusion-reconstruction paradigm-based methods usually fuse multiple burst images according to their similarity to a reference image, following their alignment via the optical flow or deformable convolution. However, this fusion strategy largely relies on the reference image and is limited to exploring more information among burst images.
\textbf{\emph{2) Dataset:}} BurstSR captures multiple LR images with a smartphone in burst mode and a corresponding HR image with a DSLR camera. Thus, several unexpected issues are nontrivial: 
\emph{a)} data misalignment (even distortion) among burst LRs and their HR counterparts; 
\emph{b)} cross-device gap between LRs and HR captured by different cameras; 
and \emph{c)} unfair model evaluation on warped SR predictions by introducing GT HR. Moreover, BurstSR can be cast as a coupled task of burst image SR and enhancement.

To address these issues, we propose the Federated Burst Affinity Network (\textbf{FBAnet}), and make an attempt to build a new real-world burst image SR dataset, named \textbf{RealBSR}. 
Our RealBSR dataset is captured in quick succession a sequence of LR images and one HR image under a continuous shooting mode with the optical zoom strategy, like RealSR~\cite{RealSR}. 
It provides a real-world benchmark for image detail reconstruction of real-world burst SR applications, avoiding the color style change in terms of the original LR data, especially, burst RAW inputs have no ISP process and often, for faithful high-resolution image predictions, especially for sensitive applications, \eg, medical imaging.

Our FBAnet employs a simple-yet-effective alignment algorithm via a homography matrix from a structural and global aspect. 
Then, a Federated Affinity Fusion (FAF) module is introduced to aggregate inter- and intra-frame information through affinity difference maps, aiming to not only focus on pixels consistent with the reference frame for global content reconstruction but also highlight the distinction among frames to absorb complementary information. 
The fused representations pass through the burst representation decoding  module to integrate local features extracted by convolutions with the global long-range dependencies of self-attentions for HR image reconstruction.

In a nutshell, our contributions are summarized below:
\begin{itemize}
\vspace{-5pt}
    \item We make an effort to establish a Real-world Burst Super-Resolution benchmark, \ie, RealBSR, which has two versions consisting of RAW and RGB images. 
    RealBSR has a great potential to inspire further researches for realistic burst SR applications.
\vspace{-5pt}
    \item We propose a Federated Burst Affinity network to address real-world burst image super-resolution, which
    derives the affinity difference maps of burst images to federate inter- and intra-frame complementary information for reconstructing more image details.
\vspace{-5pt}    
    \item We have conducted extensive experiments on RAW and RGB versions of RealBSR to benchmark existing state-of-the-art methods. Empirically, the efficacy of our FBAnet has been justified with superior SR performances from quantitative and qualitative aspects.

\end{itemize}

\section{Related Work}\label{sec:related}
\subsection{Single Image Super-Resolution}

SRCNN~\cite{SRCNN} pioneers CNN to image SR, inspiring numerous follow-ups. 
Fueled by the evolving of deep neural networks~\cite{Resnet,GAN,DenseNet,Transformer}, a series of seminal SISR methods have been built to achieve significant advances, \eg, VDSR~\cite{VDSR}, EDSR~\cite{EBSR}, SRResNet~\cite{SRRESNET}, ESRGAN~\cite{ESRGAN}, DRN~\cite{DRN}, SwinIR~\cite{SWINir}, \textit{etc.}
Nevertheless, considering the over-cost collection of real-world LR-HR image pairs, those methods turn to map synthetic LR images to their HR counterparts, which is constantly criticized for poor model generalization in practical scenarios.
To facilitate the exploration of real-world image SR, great efforts have been made on building functional benchmarks, \eg, SRRAW~\cite{zoomlearn}, RealSR~\cite{RealSR}, and DRealSR~\cite{CDC},  following the optical zoom manner to capture paired LR-HR images.
Meanwhile, LPKPN~\cite{RealSR} has been proposed to employ a Laplacian-based network for non-uniform kernel estimation. 
Encountering heterogeneous image degradation, CDC~\cite{CDC} proposes a gradient-weighted loss to adapt to diverse challenges in reconstructing different regions.

\subsection{Multi-Frame Super-Resolution}
\vspace{-5pt}
With great potential to remedy the intrinsic ill-posed SISR problem, MFSR pursues absorbing authentic sub-pixel details contained in the image sequences towards real-world applications.
In the early times of MFSR, Tsai and Huang~\cite{tsai1984multiframe} contribute the first fair solution. 
Afterward, taking advantage of deep learning, TDAN~\cite{TDAN} introduces deformable convolutions to mitigate the misalignment problem between neighboring frames and the reference frames. Similarly, EDVR~\cite{EDVR} and EBSR~\cite{EBSR} build a pyramid structure facilitating the motion compensation during the alignment procedure. MFIR~\cite{MFIR} presents a deep reparametrization algorithm that transforms Maximum A Posteriori (MAP) formulation to the latent space for better reconstruction. BIPNet~\cite{BIPNet} introduces a pseudo-burst feature fusion method to allow flexible information exchange among frames. In addition, BSRT~\cite{BSRT} builds the reconstruction module based on Swin Transformer, which further improves the performance.

For real-world burst image SR, Bhat et al. \cite{DBSR} establish a dataset consisting of LR burst images captured from a smartphone and HR counterparts from a DSLR camera and introduce an encoder-decoder-based model to deal with unknown pixel-wise displacement with optical flow estimation and merge aligned frames with an attention mechanism. 

\begin{figure*}[t!]
    \centering
    \subfloat[Pixel shift computed by SIFT method according to keypoint shift]{\includegraphics[width = 0.25\textwidth]{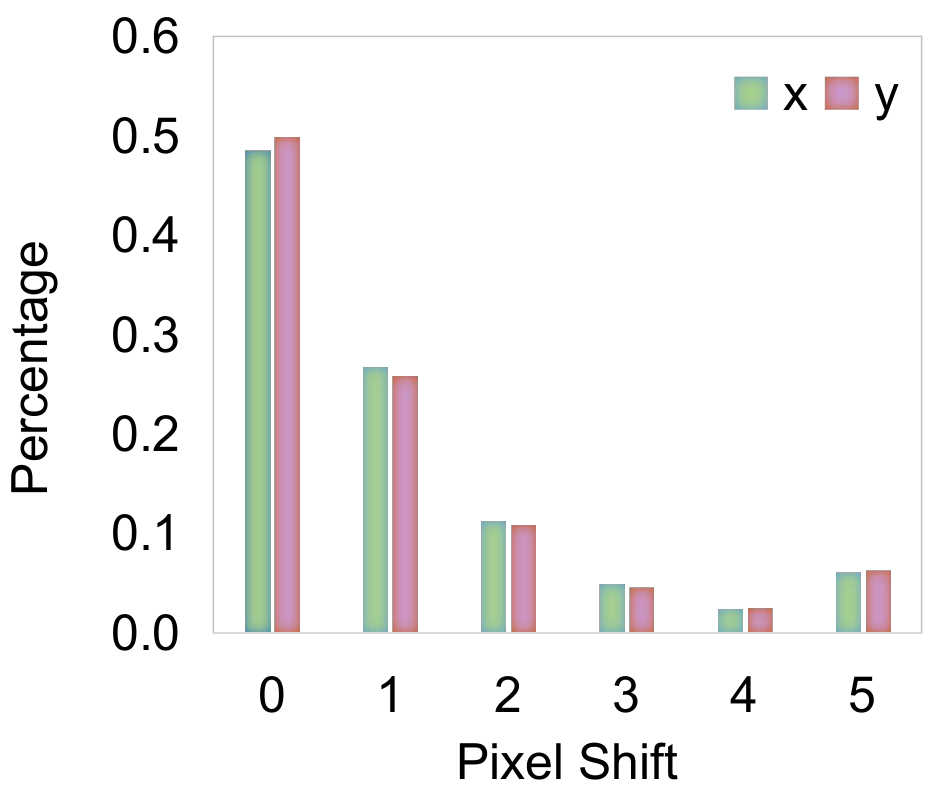}
    \label{fig:pixelshift}
    }
	\hfill
	\subfloat[Pixel shift within a range of one pixel]{\includegraphics[width = 0.35\textwidth]{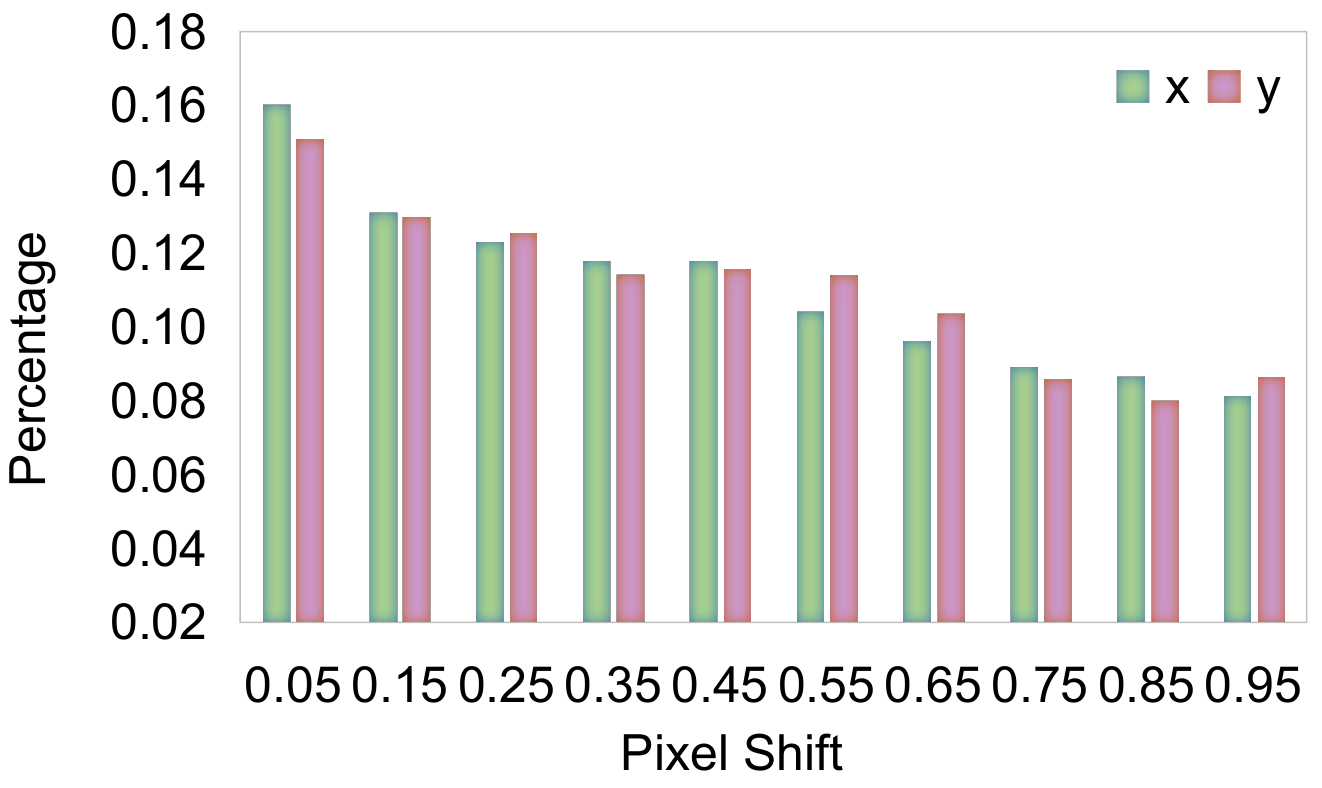}
    \label{fig:subpixelshift}
    }
	\hfill
	\subfloat[Image Diversity]{\includegraphics[width = 0.3\textwidth]{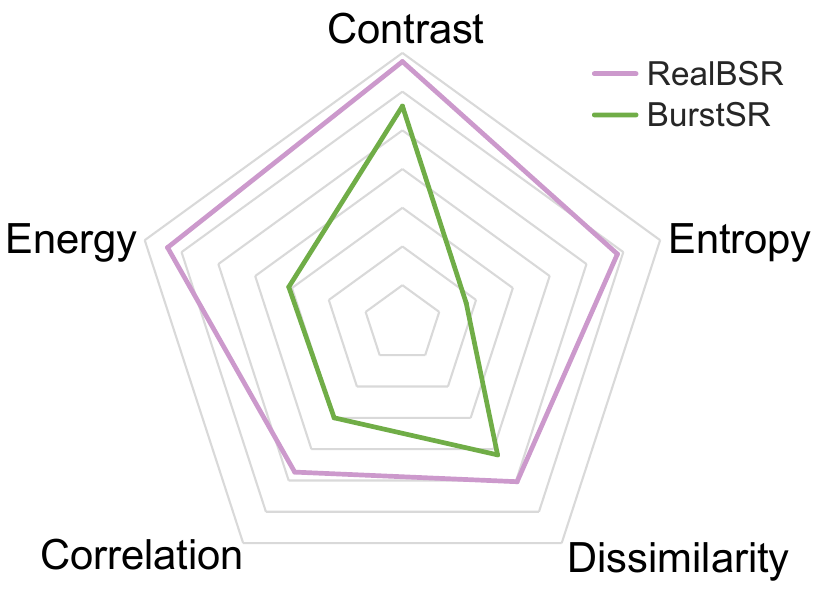}
    \label{fig:diversity}
    }
	\vspace{-8pt}
    \caption{Characteristic analyses of our RealBSR dataset.
    }
    \label{fig:DRealBSR}
    \vspace{-10pt}
\end{figure*}

One typical challenge of burst SR lies in the fusion strategy. It is common to use the affinity maps between one frame and base frame as fusion weights. However, this would be stuck in a devil: it just focuses on what is similar to the base frame for all the other frames and uses those similar pixels as complementary to the base frame; but, it is limited to excavate comprehensively complementary relations among frames. 
%
%
%
Besides, when examining the line of research works conducted on the BurstSR dataset, three issues are also worthy of being taken into serious consideration (Sec. 3), \ie, data misalignment among LRs and HRs, cross-device distribution because of different imaging cameras for capturing LRs and HRs, and unfair model evaluation with warped SR predictions via their ground-truth HRs.

In this work, we propose to leverage the federated affinity fusion strategy in our FBAnet model, comprehensively investigating complementary pixel displacements among a sequence of burst images. Meanwhile, we build a real-world burst image super-resolution dataset, named RealBSR, aiming to facilitate further exploration of real-world burst SR.

\section{RealBSR: A new Benchmark}
\label{sec:DRealBSR}

BurstSR is the only existing dataset for real-world burst image super-resolution and enhancement, which has three typical issues. 
\textit{\textbf{1) Data Misalignment.}} The distortion between LRs and their HR counterparts is distinct. It possibly results in a severe misalignment between paired LR and HR images. 
Such serious mismatches 
yield counterproductive super-resolution results with few details reconstructed from burst LR images. 
\textit{\textbf{2) Cross-Device Distribution.}} Since LR sequences and HR counterparts are captured by a smartphone and a DSLR, respectively, the difference of imaging devices would inevitably lead to a cross-device gap between them. Therefore, it has to cast this task on the BurstSR dataset as a combination of burst image super-resolution and enhancement. 
\textit{\textbf{3) Evaluation Deficiency.}} The evaluation routine for BurstSR is that a generated final SR image is warped with the reference of its ground-truth HR and then this warped SR image is used to compute the evaluation metrics with the same ground-truth HR. This is rather problematic and even not fair to truly evaluate the model performance with the aid of GTs. Besides, the calculated metric values (\eg, PSNR) cannot well reflect the visual quality, which means pursuing a higher PSNR on the BurstSR dataset is not positively related to better reconstruction quality.
This evaluation strategy greatly attributes to data misalignment and cross-device distribution, inviting a great challenge for evaluation. 

In this work, we build a real-world burst super-resolution dataset, named RealBSR. It consists of 579 groups (RAW version) and 639 groups (RGB version) for the scale factor 4. Each group has 14 burst LR images and a GT HR image. 
\begin{figure}[t!]
     \centering
     \includegraphics[width=0.49\textwidth]
    {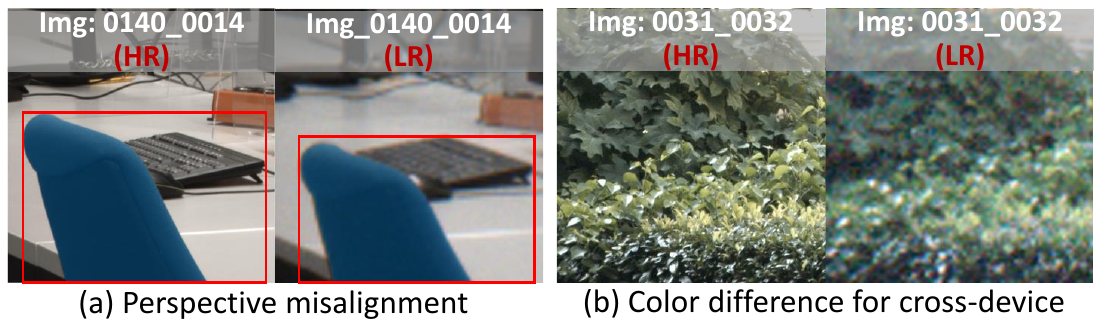}
     \vspace{-14pt}
     \caption{Limitation examples in BurstSR~\cite{DBSR}.}
     \vspace{-12pt}
     \label{fig:pipeline}
\end{figure}

\begin{figure*}[th!]
     \centering
     \includegraphics[width=1.0\textwidth]{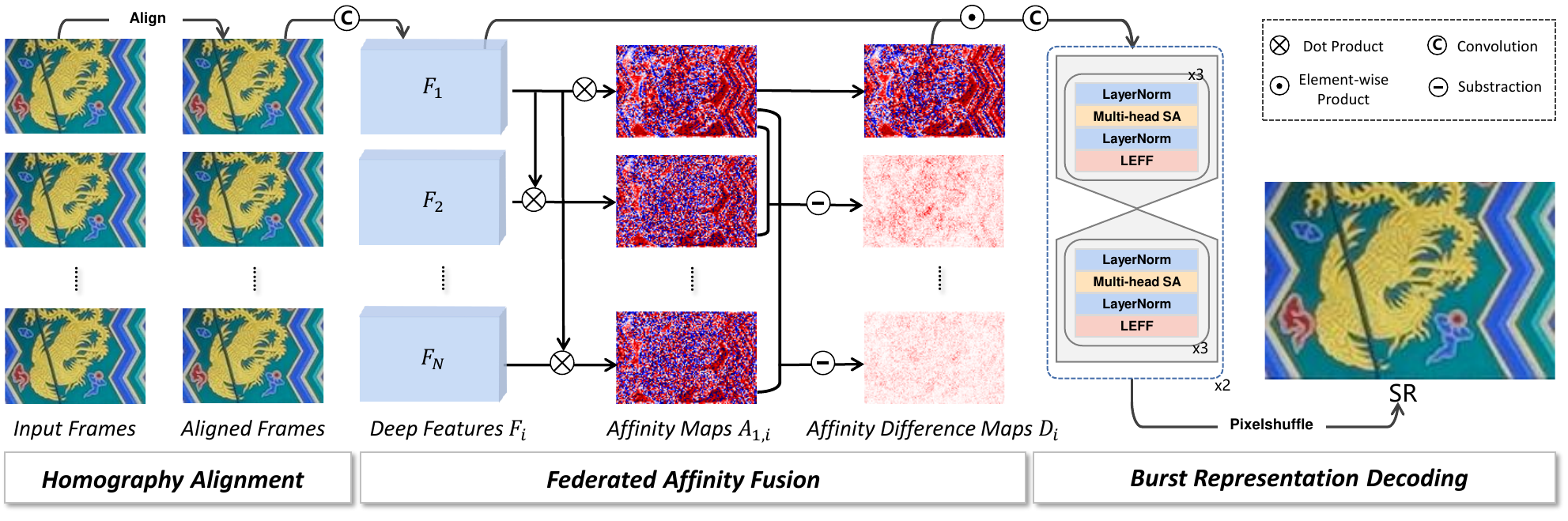}
     \vspace{-8pt}
     \caption{Workflow illustration of the proposed FBAnet, which contains three main components, including homography alignment, federated affinity fusion (\textit{cf.} Eq.~\eqref{equ:faf_fused_feature} and Eq.~\eqref{eq:guidance maps}), and burst representation decoding.}
     \vspace{-12pt}
     \label{fig:pipeline}
\end{figure*}

\subsection{Collection and Processing}
We use the optical zoom strategy for data collection, similar to RealSR~\cite{LPKPN} and DRealSR~\cite{CDC}. With a Sony DSLR camera (Alpha 7R), we capture a sequence of 14 LR images by pressing the camera shutter and optically zoom the camera to shoot an HR image. Those images are collected in various scenes, \emph{e.g.}, buildings (museum, church, office building, tower, \emph{etc.}), posters, plants/trees, sculptures, and ships. Our indoor and outdoor images have 21 and 618 groups, respectively. 
For each group of burst data, since LR and HR images have different fields of view, we employ SIFT to crop LR sequences under the reference of collected HR counterpart. 

Considering the distortion of RAW images is not addressed by the camera, their center regions are cropped into our RAW version dataset, named RealBSR-RAW. 
Besides, the RGB version of RealBSR, termed RealBSR-RGB, is also provided. Since the collected RGB images are processed by the camera ISP, it also needs color and luminance correction between LRs and their HR counterparts. 

To facilitate the model training, we crop the inputs into 160$\times$160 patches, similar to RealSR. Accordingly, the RealBSR-RAW dataset has 20,842 groups of paired patches for training and 2,377 groups for testing. Similarly, the RealBSR-RGB dataset has 19,509 groups of paired patches for training and 2,224 groups for testing.


\subsection{Characteristic Analysis}\label{sec:dataset_analysis}

\noindent
\textbf{Pixel Shift} is computed between the base frame (the first LR frame) and the other 13 frames. 
In~\cref{fig:pixelshift}, 50$\%$ offsets between frames are under 1 pixel, but 25$\%$ offsets in the range of (1,2) and 25$\%$ larger than 2 pixels, indicating the model still needs an alignment module to eliminate large offsets. Instead of other external factors like moving objects and inconsistent colors, the large pixel shifts in RealBSR are caused by intense hand tremors. In ~\cref{fig:subpixelshift}, the sub-pixel shifts are distributed evenly in the range of (0,1) which provides abundant information for SR improvement.

\noindent
\textbf{Image Diversity.}
We employ grey-level co-occurrence matrix (GLCM), which is widely used to measure image textures~\cite{GLCM}, to analyze the image diversity. With GLCM, we derive five second-order statistic features from all the training images, \ie, Haralick features~\cite{GLCM}, including image \emph{contrast}, \emph{entropy}, \emph{dissimilarity}, \emph{correlation} and \emph{energy}, ~\cref{fig:diversity}. 
\emph{Contrast} measures the intense changes between contiguous pixels, and \emph{dissimilarity} is similar to \emph{contrast} but increasing linearly. \emph{Energy} measures texture uniformity, and \emph{entropy} measures the disorder of the image, which is negatively correlated with energy. \emph{Correlation} measures the linear dependency in the image.

\vspace{-5pt}
\section{FBAnet: A New Method}\label{sec:method}
\label{sec:method}
\vspace{-5pt}
\subsection{Overview}
\vspace{-5pt}
In comparison with SISR, MFSR pursues favorable pixel-wise displacements to facilitate realistic detail reconstruction. Since it is not easy to exactly figure out the displacement association among different burst LR images, how to fuse burst images remains intractable.
What's worse, stemming from physical camera shake during imaging, it occurs more unexpected and non-uniform pixel shifts.
To address this issue, we propose a federated burst affinity network to move towards real-world burst SR by effectively integrating informatic sub-pixel details contained in multiple frames.

Our FBAnet follows a conventional alignment-to-fusion paradigm, ~\cref{fig:pipeline}. 
Formally, given an LR image sequence $\{x_i\}^N_{i=1}$ of $N$ burst observations as the input, our model will yield a high-resolution image prediction $\hat y_i$ for a scale factor $s$, where their ground-truth (GT) HR counterpart is denoted as $y_i$. 
With the randomness of pixel-wise shift among different burst images, the fused features would not be perfect enough to directly support the reconstruction of image details. 
Without loss of generality, the first frame $x_1$ is regarded as the reference frame to align the other images in the sequence by their homography matrix $H$.
Then, FBAnet employs a federated affinity fusion strategy to aggregate multiple frames and utilizes two hourglass Transformer blocks to take over the fused features for the final decoding phase of high-resolution image prediction.

\subsection{Homography Alignment}
\vspace{-5pt}
Captured in a quick succession, the pixel-wise displacements among burst images mainly stem from camera motion and scene variations, which are usually regarded to be complementary for reconstructing more details. Before fusing them, we align those images firstly, avoiding the information confusing or discrepancy and leading to blurry super-resolution predictions even with unpleasant artifacts. 

We employ a simple homography matrix~\cite{homography2001} for the alignment from a global and structural manner. Specifically, 
a 3$\times$3 homography matrix $H_t$ between $i$-th frame $x_i$ and the base frame $x_1$ indicates the transformation with respect to Correlation Coefficient Maximization (ECC) criterion~\cite{evangelidis2008parametric}. Each frame is warped with the homography matrix by taking the base frame as reference for alignment.
\subsection{Federated Affinity Fusion}\label{sec:FAF}
\vspace{-5pt}

To take full advantage of potential information, the aligned images pass through the fusion module. As fused outputs should not only be consistent with the base frame but also incorporate additional signals from other frames, we propose a Federated Affinity Fusion (FAF) module to aggregate inter-frame and intra-frame information,~\cref{fig:pipeline}. 
Our FAF determines how informative the final result could be by assigning pixel-wise fusion weights on each frame, which serves the core of the whole burst paradigm. It is noteworthy that one-order affinity maps, \ie, the differences between two affinity maps, are leveraged to determine the weights, rather than the affinity or attention maps. 
Specifically, we extract deep features $F_i$ from aligned images 
with two convolutional layers. The affinity map $A$ between two frames is the dot product of their features, \ie, $A_{i,j}=F_i \cdot F_j$. 

\begin{figure}[th!]
\centering
\subfloat[Fusion from pixels in different burst LRs. (\emph{Circle}, \emph{Triangle} and \emph{Semicircle} indicate pixels in 3 burst frames, shown in the corresponding positions of HR pixel grids.)]{\includegraphics[width=0.45\textwidth]{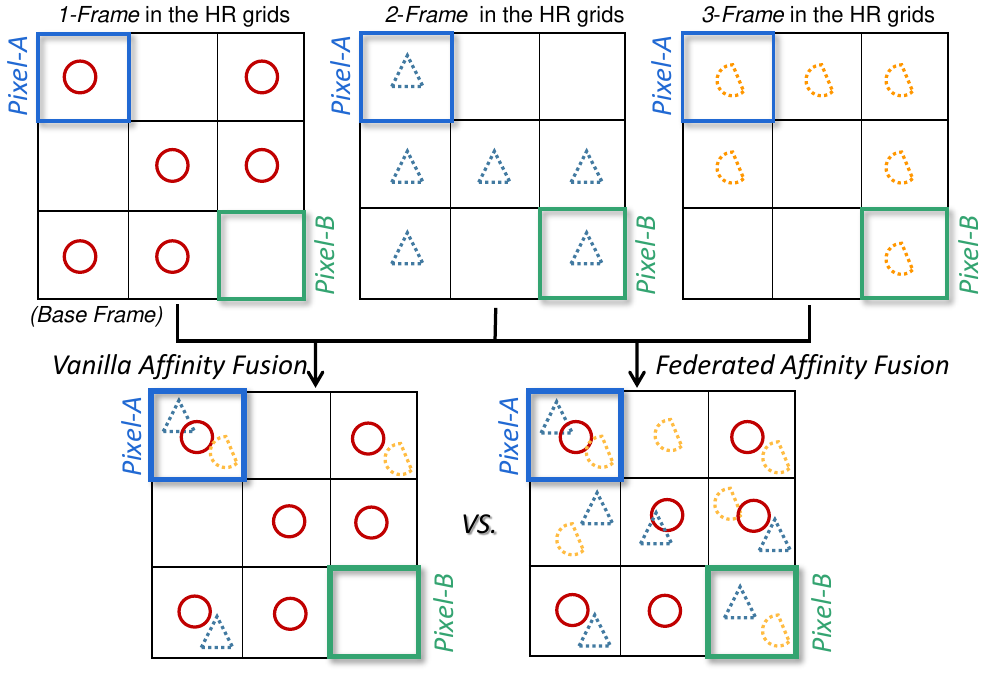}%
\label{fig:Fusion}}
 \hspace{5mm}
\subfloat[Fusion results on each of 9 HR pixels. (Height: signal intensity.)] 
{\includegraphics[width=0.48\textwidth]{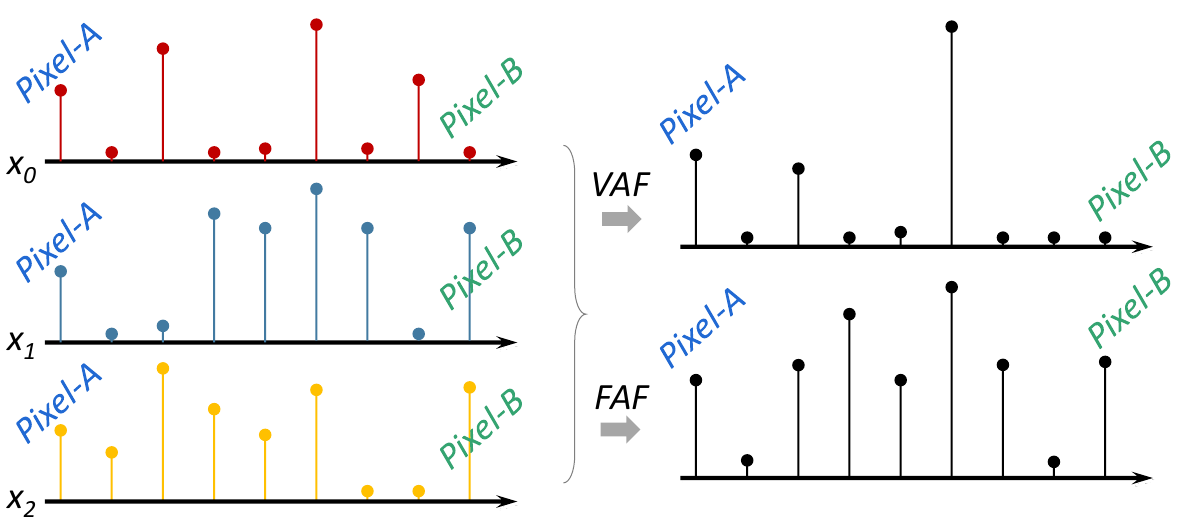}%
\label{fig:Fusion-2}}
\caption{Intuitive illustration of FAF. FAF considers more complementary details from other frames (\eg, \emph{Pixel-B} in (a)), besides those similar to base frame (\eg, \emph{Pixel-A}). (b) FAF can rectify the fusion information, avoiding negative effects from easy reconstruction regions with very high similarity and encouraging those subpixels for fusion. 
}
\label{fig:Fusion_analysis}
\end{figure}

1) \emph{Vanilla Affinity Fusion (VAF):} Following wisdom that higher similarity or affinity indicates more important pixels for fusion, it is intuitive that those affinity maps are utilized to weight each frame, which is common in existing works, \eg, TSA in EDVR~\cite{EDVR}. The fused feature map of VAF can be formulated as
$M=\sum\nolimits_{i = 1}^N {{A_{1,i}} \circ {F_i}}$, where $\circ$ is the element-wise product. 
%
As intuitively illustrated in~\cref{fig:Fusion}, 
VAF focuses on pixels from other frames consistent with the reference ones in the base frame. Consequently, information similar to the base frame would be kept (\eg, \emph{Pixel-A} in~\cref{fig:Fusion}), while important details only appear in other frames are ignored (\eg, \emph{Pixel-B} in~\cref{fig:Fusion-2}). 

2) \emph{FAF:} 
Despite the higher affinity of $x_i$ ($ \forall i \ne 1$) indicating the higher similarity to the base frame $x_1$, there also underlies two adverse effects, especially for real-world burst images:  
(a) Their easy reconstruction regions (\eg, the flat) would also have large affinity values for $x_i$, $ \forall i\ne1$, which would unexpectedly drive the model to pay more attention to those regions, resulting in over-fitting. 
(b) As for the imperfect alignment and pixel shift, even the key pixels in detail-rich regions may not have large affinity values to be highlighted for fusion. 

To address these issues, our FAF additionally considers the affinity difference maps to distinguish specific differences between one frame from other frames. 
Consequently, our FAF would pay attention to those complementary details do not appear in the base frame, \eg, \emph{Pixel-B} in~\cref{fig:Fusion}.
%
%
%
The affinity difference map of the $i$-$th$ frame can be expressed as
\vspace{-18pt}
\begin{gather}
    D_1 = A_{1,1}; ~~~~D_{i}=d(A_{1,i}, A_{1,1}), ~when~~ i \ne 1,
    \label{eq:guidance maps}
\end{gather}
\vspace{-20pt}

\noindent
where $d(\cdot)$ is the difference function. 

The final fused feature can be defined as,

\vspace{-18pt}
\begin{equation}\label{equ:faf_fused_feature}
  \begin{split}
M = \ & \sum\nolimits_{i = 1}^N {D_i \circ {F_{i}}}\\
= & \underbrace {A{}_{1,1} \circ {F_1}}_{{\rm{self-affinity \; feature}}} + \underbrace {\sum\nolimits_{i = 2}^N {d({A_{1,i}},{A_{1,1}})}  \circ {F_i}}_{{\rm{frame-specific \; feature}}}. \\
  \end{split}
\end{equation}
\vspace{-10pt}

\noindent
\textbf{Analysis:}
As \cref{equ:faf_fused_feature} indicates, the fused feature map consists of two components, \ie, (i) attentive features of the base frame based on its self-affinity, (ii) frame-specific features that are relatively independent of the base frame providing more complement from other frames. Given $D_i$ computed in the Euclidean space, \cref{equ:faf_fused_feature} can be derived as 
\vspace{-5pt}
\begin{equation}\label{equ:faf_fused_feature2}
  \begin{split}
M = \ & {A_{1,1}} \circ {F_1} + \sum\nolimits_{i = 2}^N {({F_i} - {F_1})} \cdot {F_1}\circ {F_i}.
  \end{split}
\end{equation}
\vspace{-15pt}

The second terms in \cref{equ:faf_fused_feature} \& (\ref{equ:faf_fused_feature2}) can be regarded as the combination of difference maps (${F_i}$--${F_1}$) and correlation maps ${F_1}\circ{F_i}$. 
The former would alleviate the issue that VAF encourages the fusion of redundant information similar to the base frame too much, which is too easy for reconstruction, \eg, the flat. 
The latter would alleviate the adverse effects derived from the misalignment due to large motions. Thus, FAF rectifies the fusion information to alleviate the adverse effects resulting from the large misalignment and the overfitting to the easy reconstruction of regions with high affinity, as illustrated in~\cref{fig:Fusion-2}.













3) \emph{FAF*:} Following the similar federating spirit, we can further extend this design of FAF. That is, the affinity maps and their different maps can take more complex federated interactions of frames into consideration, rather than only taking the base frame as a reference. Specifically, for $t$-$th$ frame, its affinity difference map can be compared to any other frame. Thus, $D_{i}=d(A_{1,i}, A_{1,m}), i, m \ne 1$ and the fused features is computed similar to \cref{equ:faf_fused_feature},
\vspace{-10pt}
\begin{equation*}\label{equ:faf_plus_fused_feature}
M=\sum\nolimits_{k = 1}^N {({A_{k,k}} \circ {F_k} + \sum\nolimits_{i = 1,i \ne k}^N {d({A_{k,i}},{A_{k,k}})}  \circ {F_i})}.~(4)
\end{equation*}
\vspace{-15pt}
\subsection{Burst Representation Decoding}
To aggregate global information for finer high-frequency detail reconstruction, we utilize the self-attention mechanism to model long-range pixel relations. Specifically, our FBAnet adopts a burst representation decoding module to explicitly model inter-dependencies among channels. 
This module has two cascaded blocks, shown in~\cref{fig:pipeline}. A block 
has an encoder and a decoder, both of which cascade three Locally-enhanced Window (LeWin) Transformer blocks~\cite{Uformer}. Each block has a LayerNorm, multi-head self-attention, a LayerNorm, and a Locally-Enhanced Feed-Forward (LeFF) layer~\cite{Uformer}. The module is followed by pixelshuffle~\cite{subpixel} for producing the final HR predictions.


Our training objective includes a Mean Absolute Error (MAE) loss for SR image reconstruction. 
In addition, on the RAW-version dataset, to mitigate the negative effects brought by a slight misalignment of the RAW-version dataset, we also introduce the CoBi loss~\cite{zoomlearn} to ease the training and enhance the visual quality of final results. 
While on the RGB-version dataset, we adopt the Gradient Weighted (GW) loss~\cite{CDC} for high-frequency detail reconstruction. 

\begin{figure*}
     \centering
     \begin{subfigure}[b]{0.95\textwidth}
         \centering
         \vspace{-5pt}
         \includegraphics[width=1.0\textwidth]
         {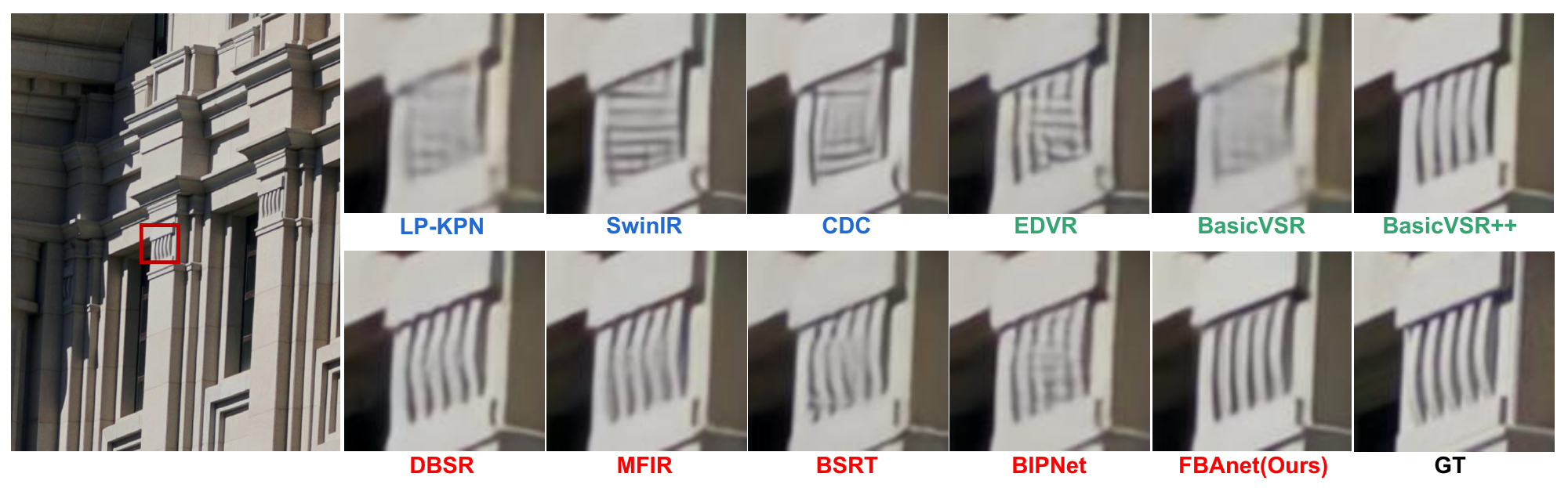}
        \vspace{-15pt}
         \caption{Burst SR results on the RealBSR-RGB dataset.}
         \label{fig:vis_comp_2}
     \end{subfigure}

     \begin{subfigure}[b]{0.95\textwidth}
         \centering
         \includegraphics[width=1.0\textwidth]{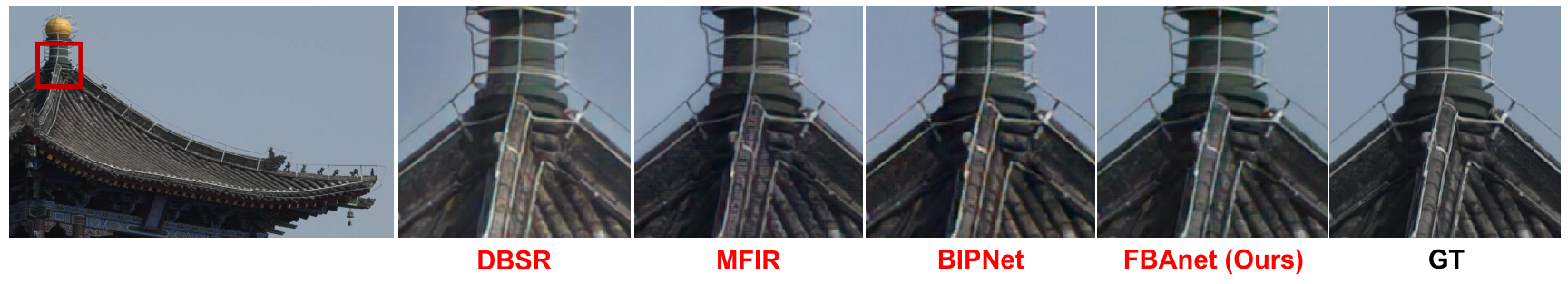}
         \vspace{-6pt}
         \caption{Burst SR results on the RealBSR-RAW dataset.}
         \label{fig:vis_comp_raw}
     \end{subfigure}
     \vspace{-5pt}
     \caption{Result visualization of existing methods including Real-world SISR (blue), video SR (green), and burst SR (red).}
     \vspace{-5pt}
     \label{fig:vis_comp}
\end{figure*}

\section{Experiments}\label{sec:exp}
\subsection{Experimental Settings}

\noindent\textbf{Datasets.} We conduct experiments on the two versions (RAW and RGB) of the proposed RealBSR benchmark at scale factor 4, real-world BurstSR~\cite{DBSR} and a synthetic burst SR dataset, SyntheticBurst~\cite{DBSR, BIPNet}, with fair comparisons. 

\noindent\textbf{Implementation Details.}
We align frames in a burst sequence using OpenCV to estimate homography matrixes, before training. 
Input images are augmented using flip and rotation in the training stage. 
The AdamW optimizer is employed and the initial learning rate is set to be 1e-4. Besides, we adopt the cosine annealing schedule to set the learning rate of each parameter group. 

\noindent\textbf{Evaluation Metric.}
On RealBSR-RAW, we adopt four evaluation metrics, \ie, PSNR, SSIM, LPIPS, and PSNR-Linear~\cite{DBSR}. The first three metrics are computed in the RGB space, and the last one is in the linear sensor space. On RealBSR-RGB, it follows the evaluation routine in the RGB image space and thus three metrics (PSNR, SSIM, LPIPS) are adopted. 
On BurstSR, the predicted SR images have to be warped by taking GT HRs as a reference before computing metrics~\cite{DBSR}, while without post-processing on our RealBSR.
%


\begin{table}
\centering
\renewcommand\arraystretch{1.3}
\resizebox{0.49\textwidth}{!}{
\begin{tabular}{|c|l|cccc|} 
\hline
Task & \multicolumn{1}{c|}{Method}  & PSNR $\uparrow$                   & SSIM $\uparrow$                  & LPIPS $\downarrow$               & PSNR-Linear $\uparrow$                \\ 
\hline
\multirow{5}{*}{\rotatebox{90}{\textit{Burst SR}}} & DBSR~\cite{DBSR}   & 20.906                            & 0.635                            & 0.134                            & 30.484                             \\
                                            & MFIR~\cite{MFIR}   & 21.562                            & 0.638                            & 0.131                            & 30.979                             \\
                                            & BSRT~\cite{BSRT}   & 22.579                            & 0.662 & \textbf{0.103}  & 30.826                             \\
                                            & BIPNet~\cite{BIPNet} & 22.896 & 0.641                            & 0.144                            & 31.311  \\
                                            & FBAnet (ours)   & \textbf{23.423}  & \textbf{0.677}  & 0.125 & \textbf{32.256}   \\
\hline
\end{tabular}}
\vspace{-5pt}
\caption{Performance comparisons on our RealBSR-RAW.}
\label{tab:Comparison_SOTA_RAW}
\end{table}

\begin{table}
\centering
\scriptsize
\renewcommand\arraystretch{1.1}
\resizebox{0.49\textwidth}{!}{
\begin{tabular}{|c|c|l|ccc|} 
\hline
\multicolumn{2}{|c|}{\multirow{1}{*}{Task}} & \multicolumn{1}{c|}{Method}                                          & PSNR $\uparrow$                   & SSIM $\uparrow$                  & LPIPS $\downarrow$                \\
\hline
\multicolumn{2}{|c|}{\multirow{3}{*}{\textit{SISR}}}                          & LP-KPN~\cite{LPKPN}     & 29.268                            & 0.863                            & 0.160                             \\
\multicolumn{2}{|c|}{}                                                        & CDC~\cite{CDC}        & 30.014                            & 0.880                            & 0.132                             \\
\multicolumn{2}{|c|}{}                                                        & SwinIR~\cite{SWINir}     & 29.924                            & 0.876                            & 0.139                             \\ 
\hline
\multirow{8}{*}{\rotatebox{90}{\textit{MFSR}}} & \multirow{5}{*}{\rotatebox{90}{\textit{Burst}~}} & DBSR~\cite{DBSR}       & 30.715                            & 0.899                            & 0.101                             \\
                                         &                                   & MFIR~\cite{MFIR}       & 30.895 & 0.899  & \textbf{0.098}   \\
                                         &                                   & BSRT~\cite{BSRT}       & 30.782                            & \textbf{0.900}                            & 0.101                             \\
                                         &                                   & BIPNet~\cite{BIPNet}     & 30.665                            & 0.892                            & 0.111                             \\
                                         &                                   & FBAnet (ours)       & \textbf{31.012}  & 0.898 & 0.102  \\ 
\cline{2-6}
                                         & \multirow{3}{*}{\rotatebox{90}{\textit{Video}~}}         & EDVR~\cite{EDVR}       & 29.708                            & 0.876                            & 0.115                             \\
                                         &                                   & BasicVSR~\cite{BasicVSR}   & 29.274                            & 0.860                            & 0.156                             \\
                                         &                                   & BasicVSR++~\cite{BasicVSR++} & 30.682                            & 0.896                            & 0.115                             \\
\hline
\end{tabular}}
\vspace{-8pt}
\caption{Performance comparisons on our RealBSR-RGB.}
\label{tab:Comparison_SOTA_RGB}
\end{table}

\begin{table}[]
\centering
\renewcommand\arraystretch{1.1}
\resizebox{0.49\textwidth}{!}{
\begin{tabular}{|c|ccc|ccc|}
\hline
\multirow{2}{*}{Method} & \multicolumn{3}{c|}{SyntheticBurst~\cite{DBSR}} & \multicolumn{3}{c|}{(real) BurstSR~\cite{DBSR}} \\
 & PSNR & SSIM & LPIPS & PSNR & SSIM & LPIPS \\ \hline
Bicubic & 36.17 & 0.909 & - & 46.29 & 0.982 & - \\
DBSR~\cite{DBSR} & 40.76 & 0.960 & - & 47.63 & 0.982 & 0.032 \\
MFIR~\cite{MFIR} & 41.56 & 0.960 & - & 48.02 & 0.984 & 0.028 \\
FBAnet (ours) & \textbf{42.23} & \textbf{0.970} & - & \textbf{48.24}  & \textbf{0.988} & \textbf{0.026} \\ \hline
\end{tabular}
}
\vspace{-8pt}
\caption{Evaluation on synthetic and real BurstSR data. '-' indicates that the LPIPS results are not provided in ~\cite{DBSR} and thus are omitted.}
\vspace{-15pt}
\label{tab:Comparison_synthetic_burst}
\end{table}

\subsection{Comparison with the State-of-the-art Methods}
We compare our model with four state-of-the-art burst SR methods, \ie, DBSR~\cite{DBSR}, MFIR~\cite{MFIR}, BSRT~\cite{BSRT}, and BIPNet~\cite{BIPNet}, and three video SR methods, including EDVR~\cite{EDVR}, BasicVSR~\cite{BasicVSR} and BasicVSR++~\cite{BasicVSR++}. 

\noindent
\textbf{Comparison with burst SR methods:} On RealBSR-RAW,~\cref{tab:Comparison_SOTA_RAW}, our FBAnet achieves the best results on PSNR, SSIM, and PSNR-Linear metrics. Notably, FBAnet improves the performance by $\sim$ 0.5dB in terms of PSNR and $\sim$ 0.945dB in terms of PSNR-Linear. Similar to the performance on RealBSR-RAW, PSNR of our FBAnet on RealBSR-RGB is also superior to other burst SR methods, validating the effectiveness of our method. ~\cref{fig:vis_comp} visualizes SR results of competing methods and ours in both RealBSR-RGB and RealBSR-RAW datasets. On RealBSR-RGB, it is clear that the state-of-the-art burst SR methods are prone to generate realistic but blurry textures, \eg, the building in~\cref{fig:vis_comp}. On RealBSR-RAW, the SR predictions of DBSR, MFIR and BIPNet have differences in image color, compared with that of our FBAnet. 
Moreover, on SyntheticBurst and real-world BurstSR, all the evaluation methods are trained from scratch in~\cref{tab:Comparison_synthetic_burst} and performance gains are also achieved by our FBAnet over existing methods.


\noindent
\textbf{Comparison with video SR methods:} To further evaluate the results of video SR methods in the real-world burst SR task, we also introduce three state-of-the-art video SR methods (\ie EDVR, BasicVSR and BasicVSR++) for comparison. Since the video SR algorithms are always based on RGB dataset, we train all these methods from scratch on RealBSR-RGB dataset. In ~\cref{tab:Comparison_SOTA_RGB}, our FBAnet outperforms the state-of-the-art video SR algorithms by $\sim$0.5dB gains (\emph{vs. BasicVSR++}) at least and 1.489dB gains (\emph{vs. EDVR}) at most. In ~\cref{fig:vis_comp_2}, it is clearly observed that visualization results of EDVR, BasicVSR and BasicVSR++ produce blurry details of the building, while our proposed FBAnet reconstructs realistic and sharp textures.

\noindent
\textbf{SISR vs. Burst SR:} To verify the benefits brought by burst SR data, we provide comparisons under the real-world SISR task. The compared methods are two representative real-world SISR methods (\ie, LP-KPN~\cite{LPKPN} and CDC~\cite{CDC}) and a Transformer-based SISR method (\ie, SwinIR~\cite{SWINir}). 
Those SISR methods only take the base frame of burst sequences as input. Compared to MFSR methods, SISR methods are characterized by generating relatively sharp and clean outputs, which could be observed from ~\cref{fig:vis_comp_2}, while suffering from the absence of informative details.

\begin{table}
\centering
\small
\renewcommand\arraystretch{1.1}
\setlength\tabcolsep{1pt}
\begin{tabular}{|cccccc|} 
\hline
Alignment & Fusion & \multicolumn{1}{c|}{Decoding} & PSNR $\uparrow$ & SSIM $\uparrow$ & LPIPS $\downarrow$ \\ \hline
\rowcolor{gray!20}
\multicolumn{6}{|c|}{\emph{Alignment}} \\ \hline
No alignment & FAF (ours) & \multicolumn{1}{c|}{ours} & 30.223 & 0.878 & 0.125 \\
Optical flow & FAF (ours) & \multicolumn{1}{c|}{ours} & 30.857 & 0.889 & 0.117 \\
Deformable & FAF (ours) & \multicolumn{1}{c|}{ours} & 30.782 & 0.891 & 0.111 \\
Homography & FAF (ours) & \multicolumn{1}{c|}{ours} & \textbf{31.012} & \textbf{0.898}  & \textbf{0.102} \\ \hline  
\rowcolor{gray!20}
\multicolumn{6}{|c|}{\emph{Fusion}} \\ \hline
Homography & VFA/TSA~\cite{EDVR} & \multicolumn{1}{c|}{ours} & 30.724 & 0.896 & 0.107 \\
Homography & FAF (ours) & \multicolumn{1}{c|}{ours} & 31.012 & 0.898 & 0.102 \\
Homography & FAF* (ours) & \multicolumn{1}{c|}{ours} & \textbf{31.197} & \textbf{0.901}  & \textbf{0.101} \\
\hline
\rowcolor{gray!20}
\multicolumn{6}{|c|}{\emph{Decoding}} \\ \hline
Homography & FAF (ours) & \multicolumn{1}{c|}{BSRT~\cite{BSRT}} & 30.890  & 0.895 &  0.106 \\
Homography & FAF (ours) & \multicolumn{1}{c|}{ours} & \textbf{31.012} & \textbf{0.898}  & \textbf{0.102} \\ \hline
\end{tabular}
\vspace{-6pt}
\caption{Evaluation about alignment, fusion, and decoding.}
\label{tab:Ablation_Study}
\end{table}

\begin{figure}[t]
     \centering
     \includegraphics[width=0.48\textwidth]{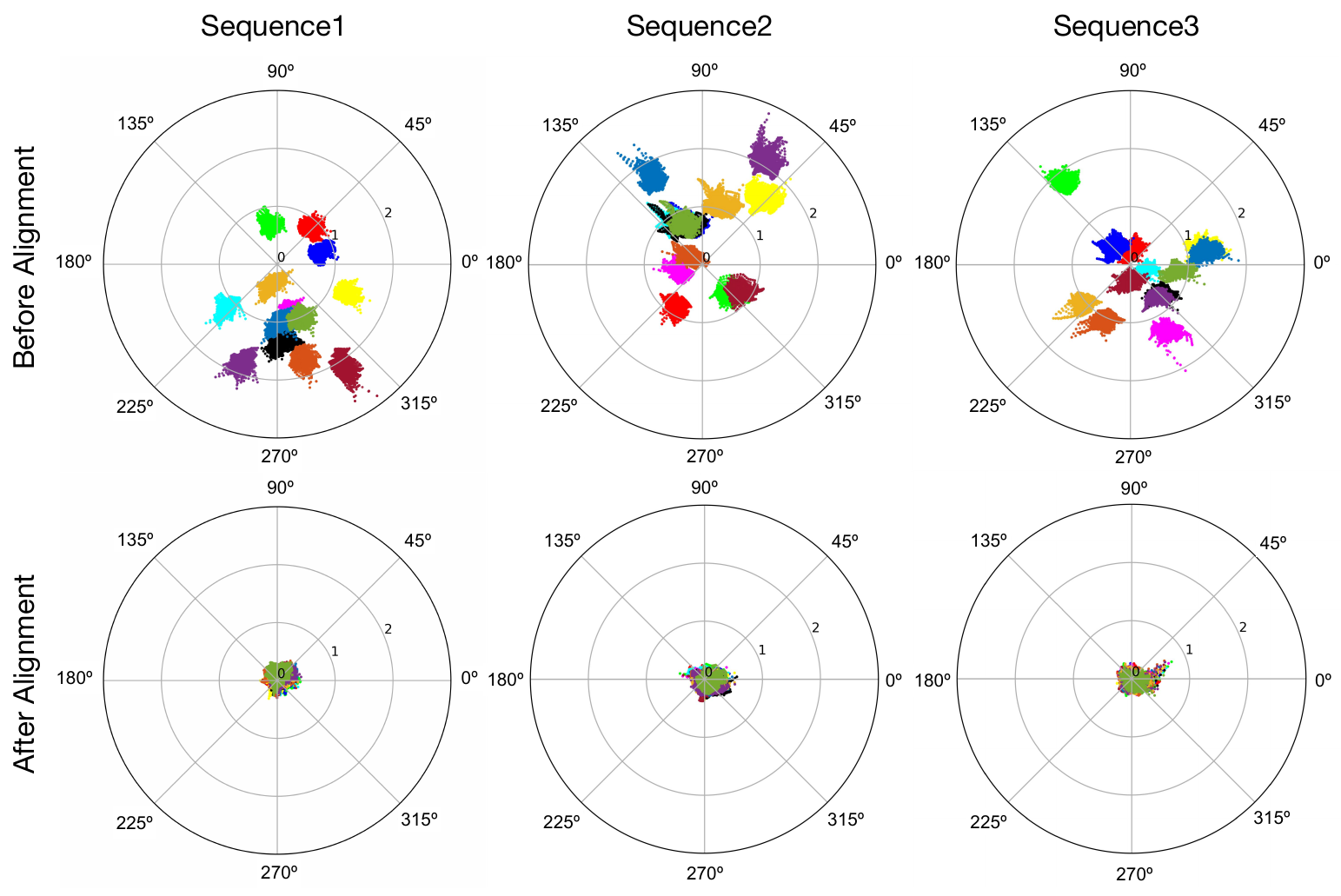}
     \vspace{-17pt}
    \caption{Three examples on pixel shift among a sequence of 14 burst images before and after our homography alignment. Each color indicates one frame in a sequence.}
     \label{fig:Shift}
\end{figure}


\subsection{Evaluation and Analysis}


\noindent
\textbf{Alignment}: We have ablatively investigated the homography alignment module and also compared it with other different alignment methods including flow-based alignment~\cite{DBSR, MFIR}  and deformable-based alignment~\cite{EDVR,TDAN}. 
As shown in ~\cref{tab:Ablation_Study}, compared with optical flow alignment~\cite{DBSR} and deformable convolutional alignment~\cite{EDVR, BSRT}, our approach outperforms them with performance improvements by 0.155dB and 0.230dB in PSNR, respectively. This demonstrates that our method is effective to align the pixel shifts in real-world burst frames, even though it is simple.

To verify our solution, we analyze motion patterns among a sequence of burst images. Taking the base frame as reference, pixel shifts of each image in a sequence are image-dependent and global-structural,~\cref{fig:Shift}. Namely, they present a relatively consistent displacement of pixels. This evidences that it is reasonable and effective to align images via homography matrix in the real-world burst super-resolution task. This is different from many existing burst algorithms that usually adopt pixel-wise alignment methods (\eg, optical flow and deformable convolutions), which rarely consider the image-wise structural motion pattern prior of the frame. 

\begin{figure}[t]
     \centering
     \includegraphics[width=0.46\textwidth]{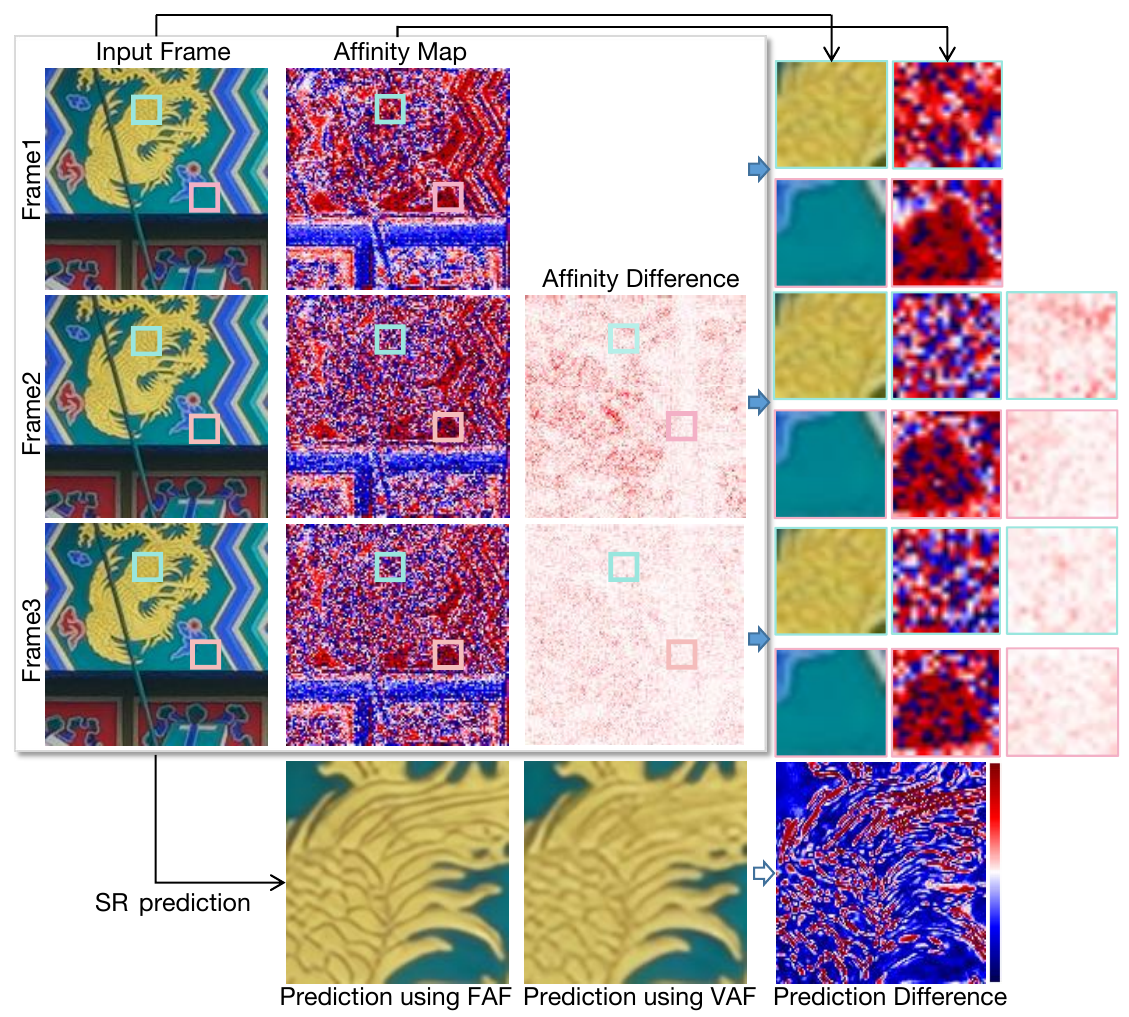}
     \vspace{-8pt}
    \caption{Visualization of affinity maps and affinity difference maps.}
     \label{fig:FAF_exp}
\end{figure}


\noindent
\textbf{Federated Affinity Fusion}: 
In ~\cref{tab:Ablation_Study}, we have evaluated the proposed federated affinity fusion module. 
In comparison with VAF using only affinity maps, our FAF introduces affinity difference maps and achieves the performance gains by 0.288dB in PSNR. And our FAF* further improves the performance by 0.185dB gains in PSNR. This indicates that our federated affinity fusion provides complementary information to the subsequent module. 

To further analyze FAF, ~\cref{fig:FAF_exp} provides the visualization of affinity maps and affinity difference maps. As discussed in~\cref{sec:FAF}, the affinity values in the flat region with few details would be rather large. Since VAF takes the affinity of one frame to base frame as fusion weight, this encourages the model to pay more attention to those easy reconstruction regions. Instead, FAF uses the affinity difference maps to lower their weights to alleviate this negative effect. 


Besides, for the difference map of Frame2 in~\cref{fig:FAF_exp}, it could be seen that the highlighted attention is different from that of Frame1 and Frame3, indicating that Frame2 also provides additional details to the fusion process. 
This can be further validated through the presented residual between HR predictions of FAF and VAF, which demonstrates that our FAF achieves better detail reconstruction than VAF, as highlighted in the prediction difference image. 

\noindent
\textbf{Burst Representation Decoding.} In~\cref{tab:Ablation_Study}, we compare our decoding module to that of BSRT with a Transformer design, under a similar architecture with the same alignment and FAF modules. Our decoding has achieved gains by 0.122dB.



\begin{table}[t]
\centering
\small
\renewcommand\arraystretch{1.1}
\setlength\tabcolsep{8pt}
\begin{tabular}{|c|ccc|} 
\hline
\multicolumn{1}{|c|}{Burst Inputs Number} & \multicolumn{1}{c}{PSNR $\uparrow$} & \multicolumn{1}{c}{SSIM $\uparrow$} & \multicolumn{1}{c|}{LPIPS $\downarrow$}  \\ 
\hline
1          & {30.139}                & {0.879}                &  {0.132}                    \\
2  &  30.616  & 0.891 & 0.113 \\
4               &              30.818                       &           0.894                          &          0.108                               \\
8               & 30.945                & 0.898                & 0.101                    \\
10              &     30.980                                &      0.899                               &                 \textbf{0.098}                        \\
14              &  \textbf{31.012}                  &  \textbf{0.898}                                   &       0.102                                 \\
\hline
\end{tabular}
\vspace{-8pt}
\caption{Evaluation on the number of burst inputs.}
\vspace{-5pt}
\label{tab:Ablation_number}
\end{table}

\vspace{-5pt}
\begin{table}
\centering
\small
\renewcommand\arraystretch{1.0}
\setlength\tabcolsep{4.5pt}
\begin{tabular}{|l|c|ccc|} 
\hline
\multicolumn{1}{|c|}{Method}    & \multicolumn{1}{c|}{Burst Inputs Data} & PSNR $\uparrow$          & SSIM $\uparrow$         & LPIPS $\downarrow$       \\ 
\hline
\multirow{2}{*}{DBSR~\cite{DBSR}}   & (base frame)$\times$14         &      29.389                    &   0.867                      &         0.150                 \\
                        &  14 burst images   &  30.715                        &     0.899                    &       0.101                   \\ 
\hline
\multirow{2}{*}{MFIR~\cite{MFIR}}   & (base frame)$\times$14         &        29.325                  &      0.865                   &         0.151                 \\
                        & 14 burst images    &         30.895                 &                0.901         &           \textbf{\textbf{0.098}}               \\ 
\hline
\multirow{2}{*}{BIPNet~\cite{BIPNet}} & (base frame)$\times$14         &      30.001                    &             0.878            &     0.136                     \\
                        & 14 burst images    &         30.665                 &         0.892                &             0.111             \\ 
\hline
\multirow{2}{*}{BSRT~\cite{BSRT}}   & (base frame)$\times$14         &      29.501                    &      0.869                   &           0.151               \\
                        & 14 burst images    &        30.695                  &       0.897                  &         0.105                 \\ 
\hline
{FBAnet}   & (base frame)$\times$14         & 30.086                  &   0.868               &   0.152                    \\
                    (Ours)    & 14 burst images    & \textbf{\textbf{31.012}} & \textbf{\textbf{0.898}} & 0.102  \\
\hline
\end{tabular}
\vspace{-8pt}
\caption{Evaluating the burst inputs' complementary content.}
\vspace{-12pt}
\label{tab:Ablation_content}
\end{table}

\vspace{5pt}
\noindent 
\textbf{The Number of Burst Image Inputs.} We investigate the impact of different numbers of burst images in a sequence and compare it with a single-frame baseline. 
And all the training processes are based on our proposed architecture. \cref{tab:Ablation_number} reveals that there has been a giant gap in the performance between the single-image baseline and multi-frame restoration results. 
Specifically, with the burst size increasing from 2 to 14, the performance also experiences a marked rise from 30.616dB to 31.197dB, which tends to be relatively saturated with the input number being close to 1. 

\noindent 
\textbf{The Complementary Content of Burst Image Inputs.} To verify the influence of contents in burst frames, we compare models trained on 14 shifted frames with models trained on 14 identical images (\ie the base frame and its 13 copies), the results of which are reported in ~\cref{tab:Ablation_content}. For the five models (\ie DBSR, MFIR, BIPNet, BSRT, and Ours) adopted, the performance gains among (base frame)$\times$14 and 14 burst images range from 0.664dB to 1.194dB, which proves the necessity and effectiveness of complementary information provided by sub-pixel information among shifted frames.

\section{Conclusions, Limitations, and Future Work}
We release a real-world burst image super-resolution dataset, named RealBSR, which is expected to facilitate exploring the reconstruction of more image details from multiple frames for realistic applications, and a Federated Burst Affinity network (FBAnet), targeting addressing the fusing issue of burst images. 
Specifically, our FBAnet employs simple homography alignment from a structural geometry aspect, evidenced by the relatively consistent pixel shift for a sequence of burst images. 
Then, a Federated Affinity Fusion (FAF) strategy is proposed to aggregate the complementary information among frames. 
Extensive experiments on RealBSR-RAW and RealBSR-RGB datasets with improved performance have justified the superiority of our FBAnet.
%

\textbf{Limitations and future work}: Our FBAnet employs a simple homography alignment. But it is not easy to extend to the video SR task with large motions, which will be addressed in our future work.  
Since noise is inevitable, addressing real-world burst super-resolution and denoising at the same time is more practical. We will be devoted to this real-world benchmark and the solutions in future work.

\section*{Acknowledgements}
This work was supported in part by National Natural Science Foundation of China (NSFC) under Grant No. U21A20470, and National Key R\&D Program of China under Grant No. 2021ZD0111601. 
Thanks a lot for the valuable help from Xiaoxiao Sun.


{\small
\bibliographystyle{ieee_fullname}
\bibliography{egbib}

\begin{thebibliography}{10}\itemsep=-1pt

\bibitem{DBSR}
Goutam Bhat, Martin Danelljan, Luc Van~Gool, and Radu Timofte.
\newblock Deep burst super-resolution.
\newblock In {\em Proceedings of the IEEE/CVF Conference on Computer Vision and Pattern Recognition}, pages 9209--9218, 2021.

\bibitem{MFIR}
Goutam Bhat, Martin Danelljan, Fisher Yu, Luc Van~Gool, and Radu Timofte.
\newblock Deep reparametrization of multi-frame super-resolution and denoising.
\newblock In {\em Proceedings of the IEEE/CVF International Conference on Computer Vision}, pages 2460--2470, 2021.

\bibitem{RealSR}
Jianrui Cai, Hui Zeng, Hongwei Yong, Zisheng Cao, and Lei Zhang.
\newblock Toward real-world single image super-resolution: A new benchmark and a new model.
\newblock In {\em International Conference on Computer Vision}, 2019.

\bibitem{LPKPN}
Jianrui Cai, Hui Zeng, Hongwei Yong, Zisheng Cao, and Lei Zhang.
\newblock Toward real-world single image super-resolution: A new benchmark and a new model.
\newblock In {\em Proceedings of the IEEE/CVF International Conference on Computer Vision}, pages 3086--3095, 2019.

\bibitem{BasicVSR}
Kelvin~CK Chan, Xintao Wang, Ke Yu, Chao Dong, and Chen~Change Loy.
\newblock Basicvsr: The search for essential components in video super-resolution and beyond.
\newblock In {\em Proceedings of the IEEE/CVF Conference on Computer Vision and Pattern Recognition}, pages 4947--4956, 2021.

\bibitem{BasicVSR++}
Kelvin~CK Chan, Shangchen Zhou, Xiangyu Xu, and Chen~Change Loy.
\newblock Basicvsr++: Improving video super-resolution with enhanced propagation and alignment.
\newblock In {\em Proceedings of the IEEE/CVF Conference on Computer Vision and Pattern Recognition}, pages 5972--5981, 2022.

\bibitem{SRCNN}
Chao Dong, Chen~Change Loy, Kaiming He, and Xiaoou Tang.
\newblock Learning a deep convolutional network for image super-resolution.
\newblock In {\em European conference on computer vision}, pages 184--199. Springer, 2014.

\bibitem{BIPNet}
Akshay Dudhane, Syed~Waqas Zamir, Salman Khan, Fahad~Shahbaz Khan, and Ming-Hsuan Yang.
\newblock Burst image restoration and enhancement.
\newblock In {\em Proceedings of the IEEE/CVF Conference on Computer Vision and Pattern Recognition}, pages 5759--5768, 2022.

\bibitem{evangelidis2008parametric}
Georgios~D Evangelidis and Emmanouil~Z Psarakis.
\newblock Parametric image alignment using enhanced correlation coefficient maximization.
\newblock {\em IEEE transactions on pattern analysis and machine intelligence}, 30(10):1858--1865, 2008.

\bibitem{GAN}
Ian Goodfellow, Jean Pouget-Abadie, Mehdi Mirza, Bing Xu, David Warde-Farley, Sherjil Ozair, Aaron Courville, and Yoshua Bengio.
\newblock Generative adversarial networks.
\newblock {\em Communications of the ACM}, 63(11):139--144, 2020.

\bibitem{DRN}
Yong Guo, Jian Chen, Jingdong Wang, Qi Chen, Jiezhang Cao, Zeshuai Deng, Yanwu Xu, and Mingkui Tan.
\newblock Closed-loop matters: Dual regression networks for single image super-resolution.
\newblock In {\em Proceedings of the IEEE/CVF conference on computer vision and pattern recognition}, pages 5407--5416, 2020.

\bibitem{GLCM}
Robert~M Haralick, Karthikeyan Shanmugam, and Its'~Hak Dinstein.
\newblock Textural features for image classification.
\newblock {\em IEEE Transactions on systems, man, and cybernetics}, (6):610--621, 1973.

\bibitem{Resnet}
Kaiming He, Xiangyu Zhang, Shaoqing Ren, and Jian Sun.
\newblock Deep residual learning for image recognition.
\newblock In {\em Proceedings of the IEEE Conference on Computer Vision and Pattern Recognition (CVPR)}, June 2016.

\bibitem{DenseNet}
Gao Huang, Zhuang Liu, Laurens Van Der~Maaten, and Kilian~Q Weinberger.
\newblock Densely connected convolutional networks.
\newblock In {\em Proceedings of the IEEE conference on computer vision and pattern recognition}, pages 4700--4708, 2017.

\bibitem{VDSR}
Jiwon Kim, Jung~Kwon Lee, and Kyoung~Mu Lee.
\newblock Accurate image super-resolution using very deep convolutional networks.
\newblock In {\em Proceedings of the IEEE conference on computer vision and pattern recognition}, pages 1646--1654, 2016.

\bibitem{SRGAN}
Christian Ledig, Lucas Theis, Ferenc Husz{\'a}r, Jose Caballero, Andrew Cunningham, Alejandro Acosta, Andrew Aitken, Alykhan Tejani, Johannes Totz, Zehan Wang, et~al.
\newblock Photo-realistic single image super-resolution using a generative adversarial network.
\newblock In {\em Proceedings of the IEEE conference on computer vision and pattern recognition}, pages 4681--4690, 2017.

\bibitem{SRRESNET}
Christian Ledig, Lucas Theis, Ferenc Huszar, Jose Caballero, Andrew Cunningham, Alejandro Acosta, Andrew~P. Aitken, Alykhan Tejani, Johannes Totz, Zehan Wang, and Wenzhe Shi.
\newblock Photo-realistic single image super-resolution using a generative adversarial network.
\newblock In {\em {IEEE} Conference on Computer Vision and Pattern Recognition}, pages 105--114, 2017.

\bibitem{SWINir}
Jingyun Liang, Jiezhang Cao, Guolei Sun, Kai Zhang, Luc Van~Gool, and Radu Timofte.
\newblock Swinir: Image restoration using swin transformer.
\newblock In {\em Proceedings of the IEEE/CVF International Conference on Computer Vision}, pages 1833--1844, 2021.

\bibitem{EDSR}
Bee Lim, Sanghyun Son, Heewon Kim, Seungjun Nah, and Kyoung~Mu Lee.
\newblock Enhanced deep residual networks for single image super-resolution.
\newblock In {\em {IEEE} Conference on Computer Vision and Pattern Recognition Workshops}, pages 1132--1140, 2017.

\bibitem{BSRT}
Ziwei Luo, Youwei Li, Shen Cheng, Lei Yu, Qi Wu, Zhihong Wen, Haoqiang Fan, Jian Sun, and Shuaicheng Liu.
\newblock Bsrt: Improving burst super-resolution with swin transformer and flow-guided deformable alignment.
\newblock In {\em Proceedings of the IEEE/CVF Conference on Computer Vision and Pattern Recognition}, pages 998--1008, 2022.

\bibitem{EBSR}
Ziwei Luo, Lei Yu, Xuan Mo, Youwei Li, Lanpeng Jia, Haoqiang Fan, Jian Sun, and Shuaicheng Liu.
\newblock Ebsr: Feature enhanced burst super-resolution with deformable alignment.
\newblock In {\em Proceedings of the IEEE/CVF Conference on Computer Vision and Pattern Recognition}, pages 471--478, 2021.

\bibitem{subpixel}
Wenzhe Shi, Jose Caballero, Ferenc Husz{\'a}r, Johannes Totz, Andrew~P Aitken, Rob Bishop, Daniel Rueckert, and Zehan Wang.
\newblock Real-time single image and video super-resolution using an efficient sub-pixel convolutional neural network.
\newblock In {\em Proceedings of the IEEE conference on computer vision and pattern recognition}, pages 1874--1883, 2016.

\bibitem{homography2001}
George Stockman and Linda~G Shapiro.
\newblock {\em Computer vision}.
\newblock Prentice Hall PTR, 2001.

\bibitem{TDAN}
Yapeng Tian, Yulun Zhang, Yun Fu, and Chenliang Xu.
\newblock Tdan: Temporally-deformable alignment network for video super-resolution.
\newblock In {\em Proceedings of the IEEE/CVF Conference on Computer Vision and Pattern Recognition}, pages 3360--3369, 2020.

\bibitem{tsai1984multiframe}
R Tsai.
\newblock Multiframe image restoration and registration.
\newblock {\em Advance Computer Visual and Image Processing}, 1:317--339, 1984.

\bibitem{Transformer}
Ashish Vaswani, Noam Shazeer, Niki Parmar, Jakob Uszkoreit, Llion Jones, Aidan~N Gomez, {\L}ukasz Kaiser, and Illia Polosukhin.
\newblock Attention is all you need.
\newblock {\em Advances in neural information processing systems}, 30, 2017.

\bibitem{EDVR}
Xintao Wang, Kelvin~CK Chan, Ke Yu, Chao Dong, and Chen Change~Loy.
\newblock Edvr: Video restoration with enhanced deformable convolutional networks.
\newblock In {\em Proceedings of the IEEE/CVF Conference on Computer Vision and Pattern Recognition Workshops}, pages 0--0, 2019.

\bibitem{ESRGAN}
Xintao Wang, Ke Yu, Shixiang Wu, Jinjin Gu, Yihao Liu, Chao Dong, Yu Qiao, and Chen Change~Loy.
\newblock Esrgan: Enhanced super-resolution generative adversarial networks.
\newblock In {\em Proceedings of the European Conference on Computer Vision}, pages 0--0, 2018.

\bibitem{Uformer}
Zhendong Wang, Xiaodong Cun, Jianmin Bao, Wengang Zhou, Jianzhuang Liu, and Houqiang Li.
\newblock Uformer: A general u-shaped transformer for image restoration.
\newblock In {\em Proceedings of the IEEE/CVF Conference on Computer Vision and Pattern Recognition}, pages 17683--17693, 2022.

\bibitem{CDC}
Pengxu Wei, Ziwei Xie, Hannan Lu, Zongyuan Zhan, Qixiang Ye, Wangmeng Zuo, and Liang Lin.
\newblock Component divide-and-conquer for real-world image super-resolution.
\newblock In {\em European Conference on Computer Vision}, pages 101--117. Springer, 2020.

\bibitem{Handheld}
Bartlomiej Wronski, Ignacio Garcia-Dorado, Manfred Ernst, Damien Kelly, Michael Krainin, Chia-Kai Liang, Marc Levoy, and Peyman Milanfar.
\newblock Handheld multi-frame super-resolution.
\newblock {\em ACM Transactions on Graphics (TOG)}, 38(4):1--18, 2019.

\bibitem{zoomlearn}
Xuaner Zhang, Qifeng Chen, Ren Ng, and Vladlen Koltun.
\newblock Zoom to learn, learn to zoom.
\newblock In {\em Proceedings of the IEEE Conference on Computer Vision and Pattern Recognition}, pages 3762--3770, 2019.

\bibitem{RDN}
Yulun Zhang, Yapeng Tian, Yu Kong, Bineng Zhong, and Yun Fu.
\newblock Residual dense network for image super-resolution.
\newblock In {\em Proceedings of the IEEE Conference on Computer Vision and Pattern Recognition (CVPR)}, June 2018.

\end{thebibliography}
}

\end{document}


\title{Towards Real-World Burst Image Super-Resolution: Benchmark and Method}

\author{
    \IEEEauthorblockN{
    Pengxu Wei$^1$ \space
    Yujing Sun$^2$ \space
    Xingbei Guo$^1$ \space
    Chang Liu$^3$ \space
    Guanbin Li$^1$ \space
    Jie Chen$^2$ \space
    Xiangyang Ji$^3$ \space
    Liang Lin$^1$ 
    } \\
    \IEEEauthorblockA{
    $^1$Sun Yat-sen University \space
    \\
    $^2$School of Electronic and Computer Engineering, Peking University, Shenzhen, China \space
    \\
    $^3$Tsinghua University
    } 
    \\
    \IEEEauthorblockA{\small{\{weipx3,guoxb7,liguanbin\}@mail.sysu.edu.cn,  yujingsun1999@gmail.com, } 
    \\     \IEEEauthorblockA{\small\{liuchang2022,xyji\}@tsinghua.edu.cn, chenj@pcl.ac.cn, linliang@ieee.org}}
}

\maketitle

This supplementary file provides additional dataset analysis and more qualitative visualizations of burst Super-Resolution (SR) results.

\section{Additional Dataset Analysis}\label{sec:supp_dataset}
\begin{figure}[b]
     \centering
     \includegraphics[width=1.02\textwidth]{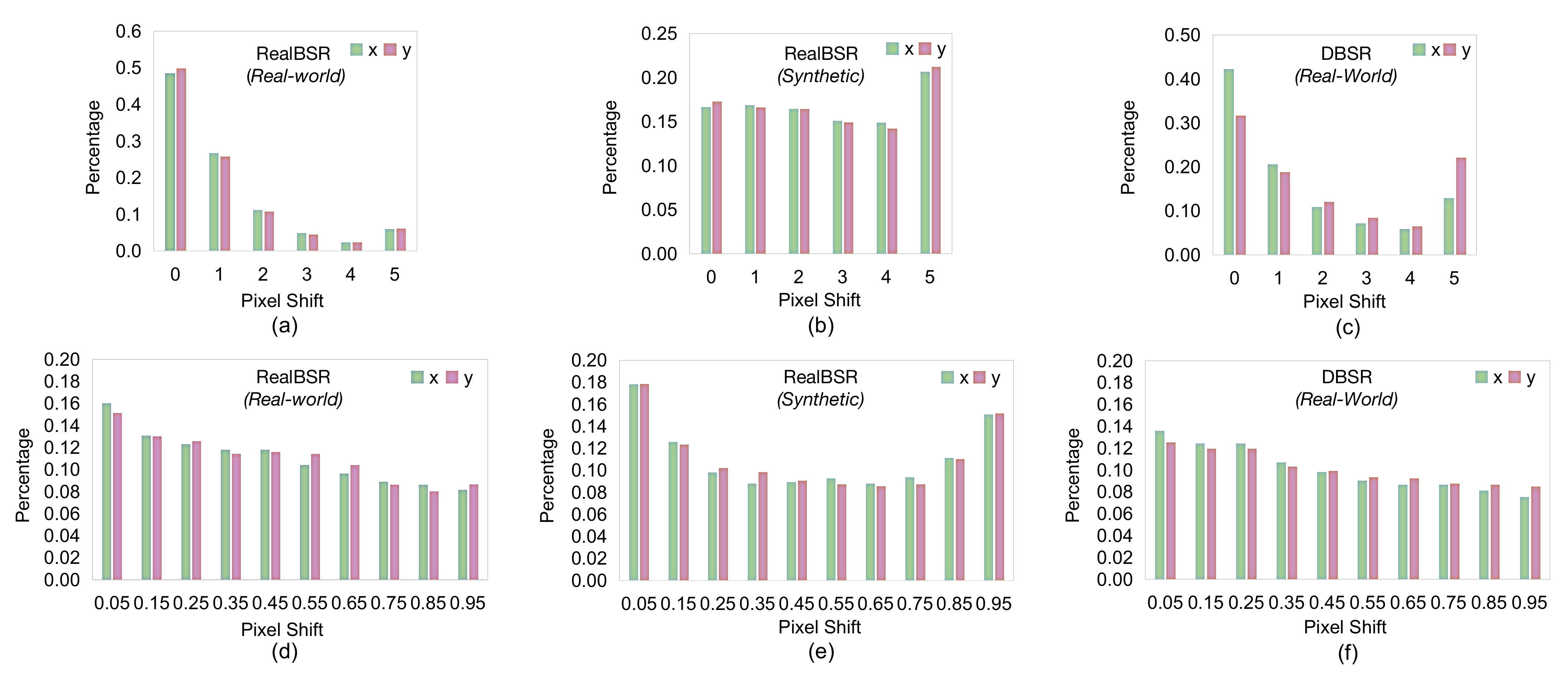}
     \caption{Pixel shifts ($a \sim c$) and sub-pixel shifts ($d \sim f$) in burst datasets. We compare our RealBSR ($a$ and $d$) with its synthetic burst version ($b$ and $e$) and real-world BurstSR dataset ($c$ and $f$).}
     \vspace{-10pt}
     \label{fig:supp_pixel_shift}
\end{figure}

\noindent
\textbf{Pixel Shift in Burst Datasets.}

\emph{Real-world vs. Synthetic.} Due to the difficulty of collecting paired real data, Multi-frame Super-Resolution (MFSR) has been limited by synthetic SR data. To figure out the differences between synthetic and real-world data, we conduct an empirical comparison on our RealBSR. A synthetic burst SR dataset, denoted as RealBSR (synthetic), is hand-crafted. Its pixel shift and subpixel shift are shown in \cref{fig:supp_pixel_shift} (b) and (e). It is observed that it has a relatively uniform distribution of pixel shift. In particular, its pixel shift in the range of (0,1) has a low percentage. The main reason is possible that the hand-craft manner is hard to simulate the real-world subpixel shift. On the contrary, real-world datasets, including DBSR and our RealBSR, have a high percentage of subpixel shifts (\cref{fig:supp_pixel_shift} (a) and (c)) and their subpixel distributions are similar (\cref{fig:supp_pixel_shift} (d) and (f)).

\emph{RealBSR vs. BurstSR.}
In the BurstSR dataset, its burst LR images and HR images are captured with different devices, \ie, cellphone and DSLR. Thus, it has to consider simultaneously the real-world burst image super-resolution and enhancement as a coupled task in a model. Instead, our RealBSR avoids this issue and provides a well-prepared benchmark for the research on real-world burst image super-resolution. We compare the pixel shift distribution of BurstSR and RealBSR in \cref{fig:supp_pixel_shift}. Larger pixel shifts beyond the range of (0,1) are observed in BurstSR. This does not indicate the larger difficulty of real-world burst SR in BurstSR, but it instead reflects a problematic phenomenon: BurstSR has a distinct image misalignment and image style change between LR and HR, as claimed in Sec. \textcolor{red}{3} in the main paper.


\begin{figure*}[t]
     \centering
     \includegraphics[width=1\textwidth]{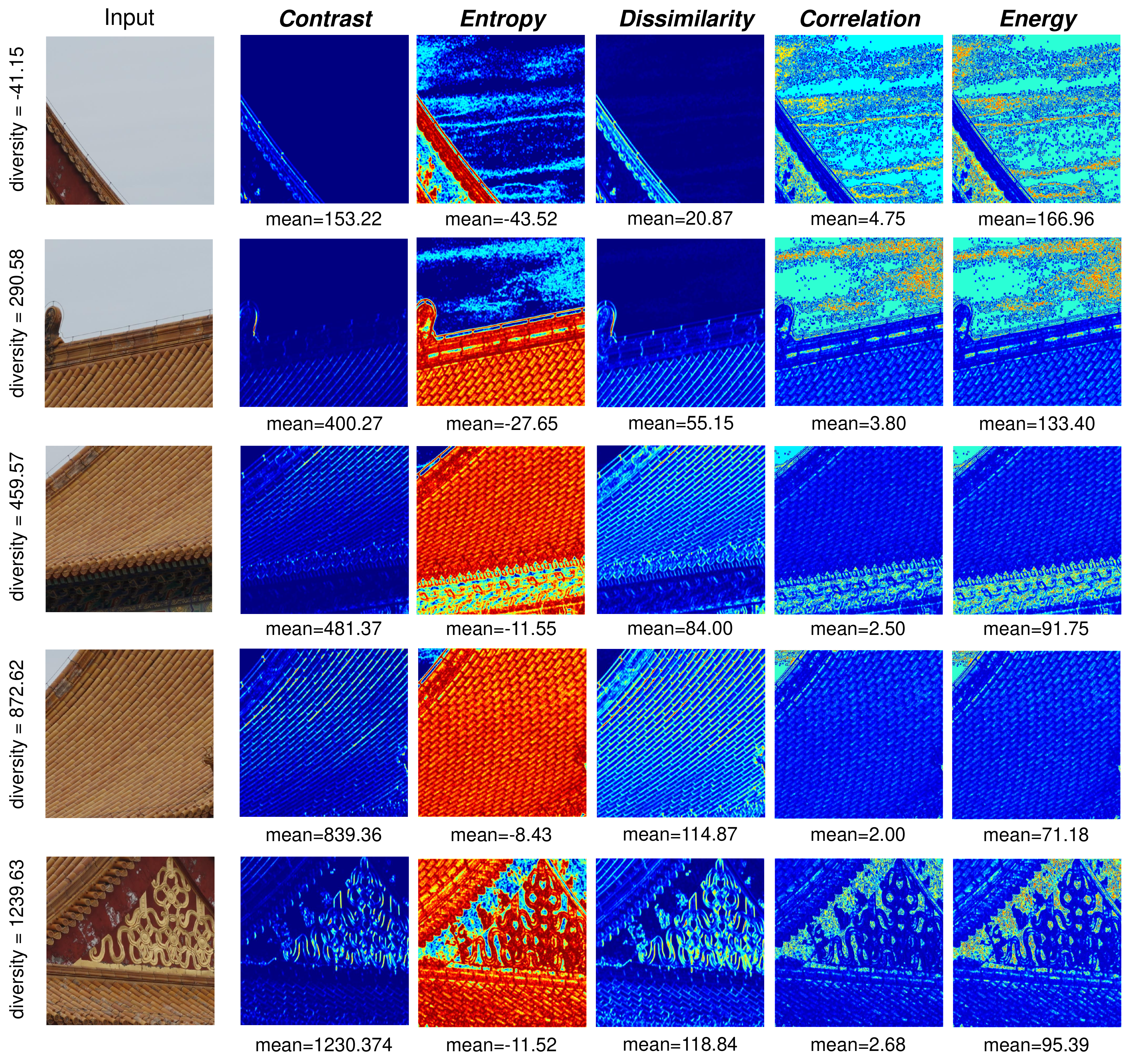}
     \caption{Visualization of image diversity analyses in terms of image \emph{contrast}, \emph{entropy}, \emph{dissimilarity}, \emph{correlation} and \emph{energy}. `\emph{mean}' indicates the average of all the elements in a feature matrix. Following \cite{moya20193d}, `\emph{diversity}' is the sum of the means of five feature matrices.}
     \vspace{-10pt}
     \label{fig:supp_glcm}
\end{figure*}

\vspace{15pt}
\noindent
\textbf{Image Diversity.}

As claimed in Sec. \textcolor{red}{3.2}, the grey-level co-occurrence matrix (GLCM) is used to analyze the image diversity~\cite{GLCM}. Accordingly, based on GLCM, we have five second-order statistic features from all the training images, \ie, Haralick features~\cite{GLCM}, including image \emph{contrast}, \emph{entropy}, \emph{dissimilarity}, \emph{correlation} and \emph{energy}. We follow their definitions of \cite{moya20193d} and provide the illustration results of images in terms of these five features, as shown in \cref{fig:supp_glcm}.
%



\vspace{5pt}
\noindent
\textbf{Model Evaluation in the BurstSR dataset.}

\begin{figure}[t]
     \centering
     \includegraphics[width=1\textwidth]{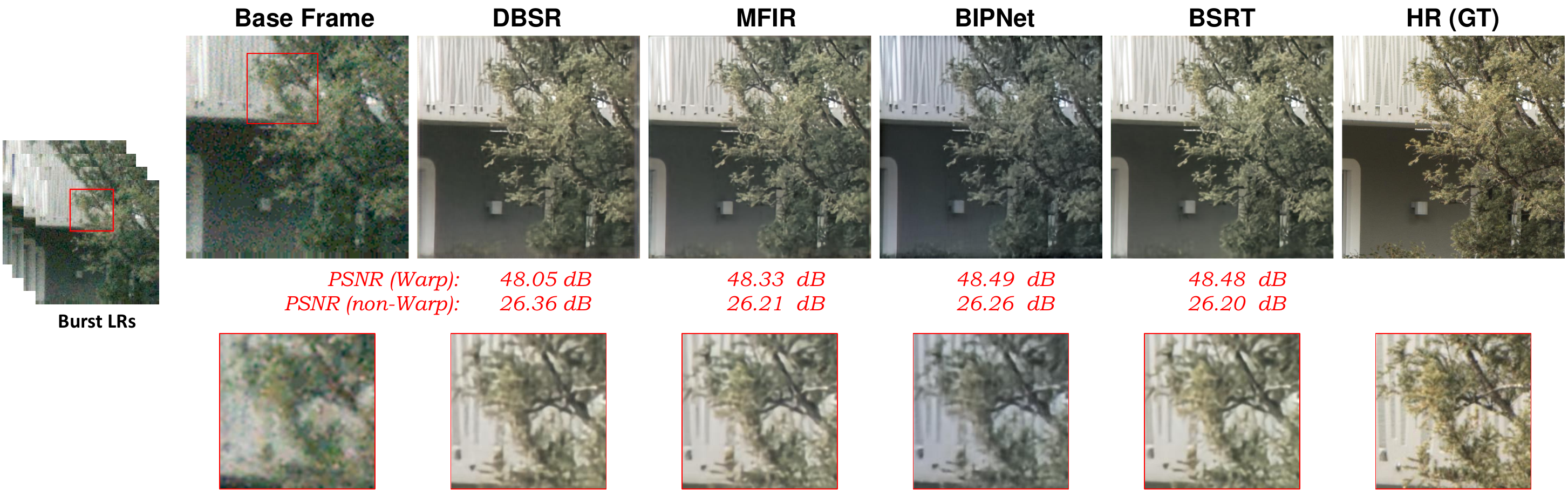}
    \caption{Model evaluation of the \emph{0120\_0009} image in the BurstSR dataset. `Warp' means that the primary SR prediction is warped with respect to the ground-truth HR image and then used for evaluation, which is evaluation manner in \cite{DBSR}. `non-Warp' means the primary SR prediction is directly used for model evaluation. It is observed that the evaluation with warped results has a very high performance, \eg, PSNR, while the SR results do not have a very high image quality. Instead, those SR prediction results have a large gap from the ground-truth HR image in terms of the image sharpness and the image color.}
    \vspace{8pt}
     \label{fig:supp_evaluation}
\end{figure}

As claimed in Sec. \textcolor{red}{3}, To evaluate models on the BurstSR dataset~\cite{DBSR}, it has an evaluation issue that the predicted SR images are firstly warped with respect to the ground-truth image and then evaluated with the ground-truth image. In \ref{fig:supp_evaluation}, we provide the SR results of different existing methods following the same evaluation strategy and in \cref{tab:supp_eva_0120}, we also present detailed quantitative results. It is observed that although models can achieve very high performance, \eg, PSNR, they have a large gap from the ground-truth HR image in terms of the image sharpness and the image color. 

\vspace{15pt}
\begin{table}[]
\centering
\begin{tabular}{|c|ccc|ccc|}
\hline
\multirow{2}{*}{Method} & \multicolumn{3}{c|}{Warp} & \multicolumn{3}{c|}{Non-Warp} \\ \cline{2-7} 
                         & PSNR    & SSIM   & LPIPS  & PSNR     & SSIM     & LPIPS   \\ \hline
DBSR                     & 48.05   & 0.984  & 0.029  & 26.36    & 0.758    & 0.108   \\
MFIR                     & 48.33   & 0.985  & 0.023  & 26.21    & 0.751    & 0.111   \\
BIPNet                   & 48.49   & 0.985  & 0.050   & 26.26    & 0.758    & 0.100     \\
BSRT                     & 48.48   & 0.985  & 0.021  & 26.20     & 0.760     & 0.110    \\ \hline
\end{tabular}
\caption{The quantitative evaluation for the \emph{0120\_0009} image in the BurstSR dataset.} 
\label{tab:supp_eva_0120}
\end{table}

{\small
\bibliographystyle{ieee_fullname}
\bibliography{egbib}
}